\documentclass[sigconf,authorversion,nonacm]{acmart}

\makeatletter
\def\@ACM@checkaffil{
    \if@ACM@instpresent\else
    \ClassWarningNoLine{\@classname}{No institution present for an affiliation}%
    \fi
    \if@ACM@citypresent\else
    \ClassWarningNoLine{\@classname}{No city present for an affiliation}%
    \fi
    \if@ACM@countrypresent\else
        \ClassWarningNoLine{\@classname}{No country present for an affiliation}%
    \fi
}
\makeatother

\usepackage{graphicx} 
\usepackage{multirow}
\usepackage{footnote}
\title{L\textsuperscript{3} Ensembles: Lifelong Learning Approach for Ensemble of Foundational Language Models\footnote{Contribution No. 1 from our lab}}

\author{Aidin Shiri$^1$, Kaushik Roy$^2$, Amit Sheth$^2$, Manas Gaur$^1$}
\affiliation{%
  \institution{$^1$University of Maryland,
  Baltimore County (UMBC), MD, USA; 
  $^2$ AI Institute, University of South Carolina, SC, USA \\
 \texttt{\{aidin.shiri, manas\}@umbc.edu},
 \texttt{kaushikr@email.sc.edu, amit@sc.edu}
  } }

\date{September 2023}

\begin{document}

\maketitle
\pagestyle{plain}

\section{Abstract}\footnote{Aidin Shiri, Ph.D. Student, UMBC}
Fine-tuning pre-trained foundational language models (FLM) for specific tasks is often impractical, especially for resource-constrained devices. This necessitates the development of a Lifelong Learning (L\textsuperscript{3}) framework that continuously adapts to a stream of Natural Language Processing (NLP) tasks efficiently. We propose an approach that focuses on extracting meaningful representations from unseen data, constructing a structured knowledge base, and improving task performance incrementally. We conducted experiments on various NLP tasks to validate its effectiveness, including benchmarks like GLUE and SuperGLUE. We measured good performance across the accuracy,  training efficiency, and knowledge transfer metrics. Initial experimental results show that the proposed L\textsuperscript{3} ensemble method increases the model accuracy  4\%$\sim$36\% compared to the fine-tuned FLM. Furthermore, L\textsuperscript{3} model outperforms naive fine-tuning approaches while maintaining competitive or superior performance (up to 15.4\% increase in accuracy) compared to the state-of-the-art language model (T5) for the given task, STS benchmark.

\begin{figure}[]
    \centering
    \includegraphics[width=3.5in]{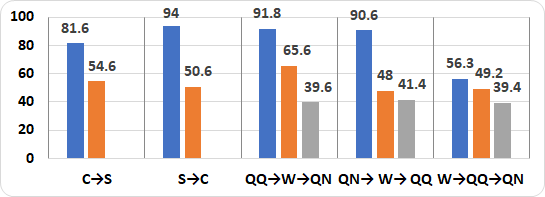}
    \caption{\footnotesize Performance (\%) of the Fine-Tuned $BERT_{base}$ on Continual Sequence of Similar GLUE Benchmark Tasks (C: CoLA, S: SST2, QQ: QQP, W: WNLI, QN: QNLI). Notation ``$C \rightarrow S$'' denotes training on CoLA and testing on SST2.  Notation ``$QQ \rightarrow W \rightarrow QN $'' indicates that the model underwent training on QQP, followed by WNLI, and was subsequently tested on QNLI.}
    \label{fig1}
\end{figure}

\begin{figure}[!t]
    \centering
    \includegraphics[width=2.5in]{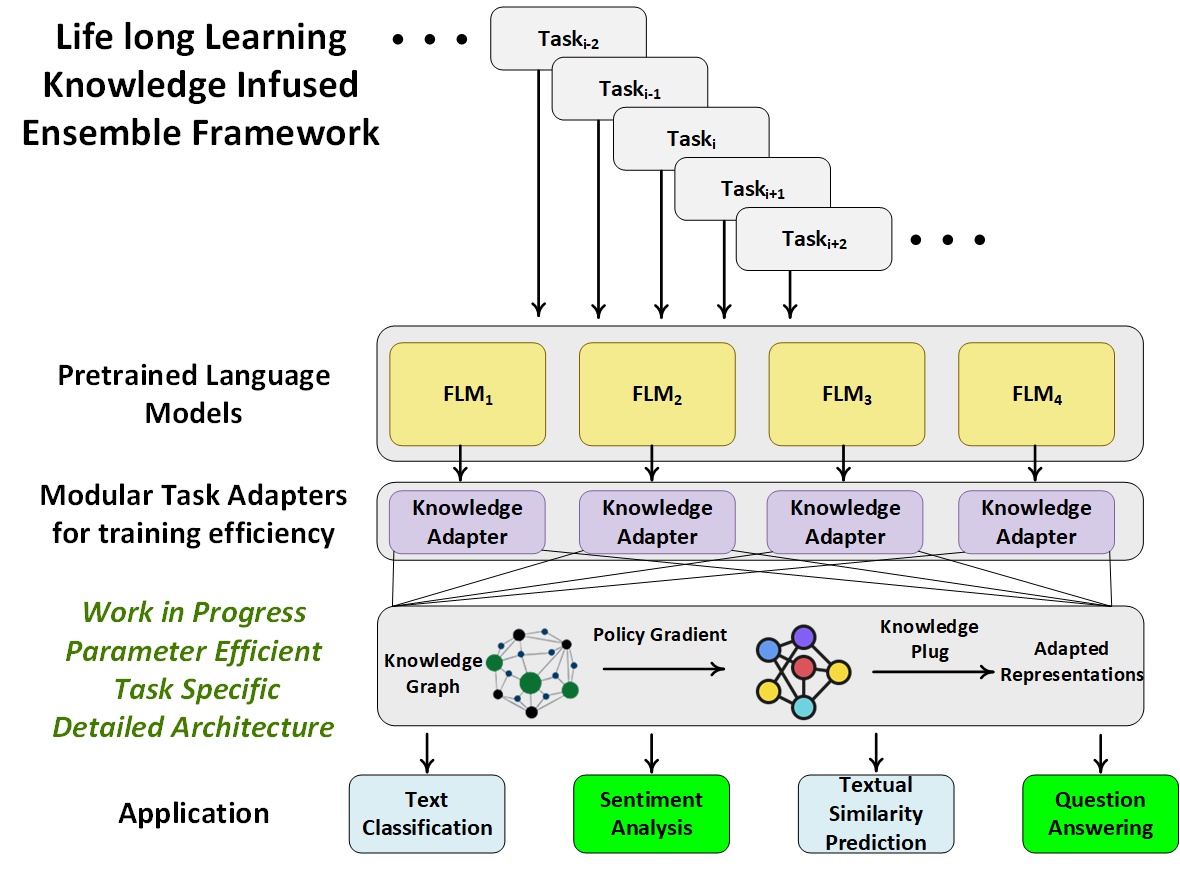}
    \caption{A framework for Lifelong Learning using Ensemble of FLMs. Blue tasks are the parts that have been implemented, and green parts are works in progress.} \vspace{-3ex}
    \label{fig33}
\end{figure}

\begin{table*}[]
\scriptsize
\renewcommand{\arraystretch}{0.7}

\begin{tabular}{cclcclc}

\toprule
\multicolumn{1}{c}{Task}                                                                 & Step & FineTuned Model                                   & \multicolumn{1}{c}{Evaluation Task} & Accuracy (A) & Comparison Baseline (CB)                                                  & Knowledge Transfer ($A-CB$)                                                                  \\
\toprule

\multirow{4}{*}{\begin{tabular}[c]{@{}c@{}} GLUE TASKS: \\ QQP and MRPC\end{tabular}}      & 1    & $BERT_{ base \rightarrow  QQP}$  & QQP                              & 91.8\%   & \begin{tabular}[c]{@{}l@{}}$BERT_{base}$: 63\%\end{tabular}   & 28.8\% (+)                                                                \\
                                                                                         & 2    & $BERT_{QQP}  $                        & MRPC                             & 71.8\%   & \begin{tabular}[c]{@{}l@{}}$BERT_{base}$:  31.6\%\end{tabular} & 40.2\% (+)                                                                \\
                                                                                         & 3    & $BERT_{ QQP \rightarrow       MRPC } $     & MRPC                             & 86.2\%   & $BERT_{  QQP }  $: 71.8\%                                                    & 14.4\% (+)                                                              \\
                                                                                         & 4    & $BERT _{ MRPC}  $                       & QQP                              & 84.9\%   & $BERT _{ QQP}  $: 91.8\%                                                    & \begin{tabular}[c]{@{}l@{}} \textbf{CF: 7\%} (-) \end{tabular}    \\  \toprule
\multirow{4}{*}{\begin{tabular}[c]{@{}c@{}} SuperGLUE TASKS:\\ BoolQ and RTE\end{tabular}} & 1    & $BERT_{base \rightarrow RTE}$ & RTE                              & 72.5     & \begin{tabular}[c]{@{}l@{}}$BERT_{base}$: 47.2\%\end{tabular} & 25.3\% (+)                                                                \\ 
                                                                                         & 2    & $BERT_{  RTE}   $                       & BoolQ                            & 60\%     & \begin{tabular}[c]{@{}l@{}}$BERT_{base}$: 37.8\%\end{tabular} & 22.2\% (+)                                                                \\
                                                                                         & 3    & $BERT_{ RTE \rightarrow  BoolQ}$      & BoolQ                            & 75.2\%   & $BERT_{RTE}  $: 60\%                                                     & 15.2\% (+)                                                                \\
                                                                                         & 4    & $BERT _{BoolQ} $ :                    & RTE                              & 43.6\%   & $BERT _{RTE}$: 72.5\%                                                        & \begin{tabular}[c]{@{}l@{}} \textbf{CF: 28.9}\% (-)  \end{tabular}

\\
\bottomrule
\end{tabular}
\caption{\footnotesize The Knowledge transfer of single FLM ``$BERT_{base}$'' on the four sample GLUE and SuperGLUE tasks. Notation ``$BERT_{ base \rightarrow  QQP}$'' denotes that the base model is fine-tuned on QQP task.}
\vspace{-3em}
\end{table*}

\begin{table*}[!t]
\scriptsize
\begin{tabular}{lcc|lccccc}
\toprule
FLM & Size     & \multicolumn{1}{l}{\begin{tabular}[c]{@{}l@{}}MSE\end{tabular}} & Ensemble & Size             & \multicolumn{1}{l}{\begin{tabular}[c]{@{}c@{}}Naïve\\ Ensemble\end{tabular}} & \multicolumn{1}{l}{\begin{tabular}[c]{@{}c@{}}Weighted \\  Ensemble\end{tabular}} & \multicolumn{1}{l}{\begin{tabular}[c]{@{}c@{}}LLM \\ Ensemble\end{tabular}} & \begin{tabular}[c]{@{}c@{}} KI \\ Ensemble\end{tabular} \\
\toprule
BERT  & 110M       & 0.39                                                                  & BERT \& DistilBERT             & 176M   & 0.382                                                                       & 0.382                                                                            & 0.393                                                                            & \textbf{0.374}                                                                \\
DistilBERT & 66M & 0.45                                                                   & BERT \& ELECTRA              & 220M   & 0.304                                                                       & 0.302                                                                            & \textbf{0.285}                                                                             & 0.288                                                                \\
ELECTRA  & 110M  & 0.34                                                                   & BERT \& RoBERTa        & 235M        & 0.294                                                                       & 0.290                                                                            & \textbf{0.275}                                                                             &   0.293                                                                \\
RoBERTa & 125M   & 0.32                                                                  & BERT \& DistilBERT   \& ELECTRA & 286M   & 0.301                                                                       & 0.316                                                                            & 0.312                                                                             & \textbf{0.295}                                                                \\
T5   & 11B      & 0.31                                                                   & BERT\& RoBERTa   \& ELECTRA  & 345M   & 0.286                                                                       & 0.282                                                                            & \textbf{0. 262}                                                                           & 0.264           \\
\bottomrule
\end{tabular}
\caption{\footnotesize Effect of Ensembling and Knowledge Infusion on the  MSE loss of the STS benchmark dataset. We see that ensembling improves performance even with models of modest size - this is especially noteworthy in the last row with the T5 model.} \vspace{-4ex}
\end{table*}

\section{Introduction}
When training models in Artificial Intelligence (AI) and Machine Learning (ML), the capacity for models to continually learn and adapt to new tasks and data distributions is a critical challenge. Typically, AI and ML models are meticulously trained on specific datasets to acquire domain knowledge, expecting that this knowledge can then be applied to perform well on unseen but related data. However, real-world applications often demand more flexibility. These applications require models to learn new tasks efficiently and retain the knowledge of previously learned tasks, thus avoiding what is known as catastrophic forgetting (CF). CF occurs when adapting a model to a new task leads to a significant loss of knowledge in previously learned tasks. 

In recent years, the research focus has shifted towards addressing this issue through the paradigm of \textbf{L}ife\textbf{L}ong \textbf{L}earning, or L\textsuperscript{3} \cite{sun2019lamol}. Existing L\textsuperscript{3} approaches have explored various strategies, including regularization, model architecture, and data-based techniques. For instance, some methods employ regularization techniques to consolidate weights associated with previous tasks when learning new ones \cite{lee2017overcoming}. Others isolate specific model parameters for different tasks or incorporate replay-based approaches using old task data to guide new task learning \cite{de2019continual}\cite{wang2020efficient}. Within the realm of Natural Language Processing (NLP), especially in resource-constrained scenarios like edge devices, there is a greater demand for straightforward and efficient Foundational Language Models (FLMs). This demand arises when comparing them to their more intricate counterparts with over a billion parameters. The practical applications of FLMs in edge devices encompass tasks like question-answering, engaging in conversations, and extracting information from visual content (e.g., named entities)\cite{sun2019lamol}\cite{gunaratna2021using}. 

\emph{This paper argues that rather than elevating the complexity of FLMs, more favorable outcomes can be attained through the fusion of multiple simpler FLMs enriched with infused knowledge.}

This paper proposes an alternative approach to L\textsuperscript{3} that leverages ensemble configurations of pre-trained models and external knowledge augmentation to combat CF without requiring compute-intensive techniques. A compelling demonstration of the concept of CF in NLP can be observed in Figure \ref{fig1}. It vividly illustrates how, when we train $BERT_{base}$ on a specific task and then assess its performance on a semantically related dataset, we witness a notable decline in its performance. Furthermore, the same phenomenon occurs when $BERT_{base}$ is trained on two similar datasets; it still experiences a decrease in performance when tested on a third dataset, even if that dataset involves a related task. This striking evidence underscores the significance of addressing CF challenges in NLP for usability in edge devices.

Traditional ensembling often involves simple or weighted aggregation methods, which can lead to suboptimal performance when solving tasks. Naive ensemble and weighted ensemble are previously proposed \cite{matena2111merging}. We introduce the Large Language Model (LLM) Ensemble and Knowledge Infused (KI) ensemble as two methods to prevent CF in FLM training or fine-tuning. In the \textit{LLM Ensemble}, we harness the power of frozen embeddings from LLMs to enhance vector representations through meticulous modulation. Specifically, we incorporate embeddings derived from ``Langchain text-embedding-ada-002 LLM" into our LLM Ensemble approach, a pivotal step in fortifying the model's vectorized representations. For the \textit{KI Ensemble}, as showcased in Figure \ref{fig33}, we capitalize on the vectorized information from Wikipedia Knowledge Graph.    



\vspace{-3ex}

\section{Experiments and Discussion}


\textbf{\textit{Tasks and Datasets:}} We employ multiple datasets from two established benchmarks in our experiments. One of these benchmarks is the General Language Understanding Evaluation (GLUE), encompassing a variety of natural language understanding tasks \cite{wang2018glue}. The second is an enhancement over GLUE called SuperGLUE, which includes a relatively more demanding and varied assortment of tasks \cite{wang2019superglue}. \textbf{\textit{Individual Model Baselines.}} As baselines, we utilize various FLMs, including BERT, RoBERTa, DistilBERT, and ELECTRA. We first assess individual FLM performance on various GLUE tasks, as illustrated in Figure 1 and Table 1.

\textbf{Knowledge Transfer and Catastrophic Forgetting of Individual Models.} We measure the individual FLM model performance on a sequence of tasks (using accuracy). Our findings indicate that while fine-tuning boosts performance on individual tasks (illustrated by base --> task in \textbf{Table 1.}), training a model fine-tuned on one task for another task causes CF of the old task (between 7\% and 28.9\%). Thus, an ensemble approach has the potential to mitigate CF while enhancing knowledge transfer.

\textbf{Ensemble Methods for Improving Performance in Resource-Constrained Systems.}
Next, we experiment with another dataset, STS, which is most representative of the tasks across the benchmark datasets. This time, we report the results of individual FLM performance vs. different ensemble configurations. We find that the ensemble performs better than the individual models, and crucially, all of the individual models are at most 500M in size - showing that ensemble methods are a good choice for resource-constrained domains (see Table 2) \cite{gunaratna2021using}.



%


\section{Conclusion and Future Work}

Fine-tuning FLMs improved task-specific performance in GLUE tasks, but transferring a fine-tuned model to another task led to a significant performance drop due to CF. Our work highlighted the need for a L\textsuperscript{3} ensemble approach to mitigate this issue, demonstrating the superior performance of ensembles over individual models. The findings emphasize the potential of ensemble methods to enhance knowledge transfer and address CF in settings where model size and efficiency are crucial for success, such as resource-constrained settings. This research will extend KI Ensemble using a reinforcement learning-based approach (see Figure \ref{fig33}, green boxes) across various knowledge-intensive NLP tasks \cite{petroni2021kilt}. 



\bibliographystyle{ACM-Reference-Format}
\bibliography{ref.bib}

\end{document}